\documentclass[letterpaper]{article}
\usepackage{proceed2e}
\usepackage[margin=1in]{geometry}
\usepackage{graphicx, float, hyperref}

\usepackage{times}

\title{Weakly-Supervised Deep Learning of \\Heat Transport via Physics Informed Loss}

\author{} 

\author{ {\bf Rishi Sharma} \\
Electrical Engineering, \\
Stanford University\\
\href{mailto:rsh@stanford.edu}{rsh@stanford.edu} \\
\And
{\bf Amir Barati Farimani}  \\
Bioengineering,\\
Stanford University
\And
{\bf Joe Gomes}  \\
Bioengineering,\\
Stanford University
\And
{\bf Peter Eastman} \\
Bioengineering,\\
Stanford University
\And
{\bf Vijay Pande} \\
Bioengineering,\\
Stanford University\\
\href{mailto:pande@stanford.edu}{pande@stanford.edu}
}

\begin{document}

\maketitle

\begin{abstract}
In typical machine learning tasks and applications, it is necessary to obtain or create large labeled datasets in order to to achieve high performance. Unfortunately, large labeled datasets are not always available and can be expensive to source, creating a bottleneck towards more widely applicable machine learning. The paradigm of weak supervision offers an alternative that allows for integration of domain-specific knowledge by enforcing constraints that a correct solution to the learning problem will obey over the output space. In this work, we explore the application of this paradigm to 2-D physical systems governed by non-linear differential equations. We demonstrate that knowledge of the partial differential equations governing a system can be encoded into the loss function of a neural network via an appropriately chosen convolutional kernel. We demonstrate this by showing that the steady-state solution to the 2-D heat equation can be learned directly from initial conditions by a convolutional neural network, in the absence of labeled training data. We also extend recent work in the progressive growing of fully convolutional networks to achieve high accuracy ($<1.5\%$ error) at multiple scales of the heat-flow problem, including at the very large scale (1024$\times$1024). Finally, we demonstrate that this method can be used to speed up exact calculation of the solution to the differential equations via finite difference.
\end{abstract}

\section{Introduction}\label{introduction}
Most of the dramatic successes of machine learning in the 21st century have utilized very large datasets in order to achieve their performance \footnote{ \href{https://www.edge.org/response-detail/26587}{https://www.edge.org/response-detail/26587} argues that the creation of high-quality datasets is a better predictor of progress in AI than algorithmic advancement.}. In supervised tasks, including in image recognition \cite{russakovsky2015imagenet}, speech recognition \cite{graves2013speech} and machine translation \cite{sutskever2014seq2seq}, large datasets had to be assembled before neural network architectures particularly suited to these applications could emerge and achieve human-level performance on these tasks. In reinforcement learning \cite{suttonbarto98RL}, successful agents, including Go champion AlphaZero \cite{silver2017alphazero} and Atari-playing DQNs \cite{mnih2015atari}, operate in easily-simulated toy environments that enable the collection of large quantities of data in the form of observations and interactions with the environment.

The paradigm of weakly-supervised learning \cite{dawn17weak-supervision} seeks to reduce the data requirements by encoding our prior knowledge into machine learning systems. In this work, we explore the ability to encode our prior knowledge about the physical world into an appropriate loss function to guide deep learning.

We are interested in inference within physical environments whose rules can be defined by differential equations. We focus on the case of 2-dimensional heat transport \cite{fourier1807heat}, whose dynamics are defined by a second-order differential equation. We seek to encode the dynamics of the system into a loss function, such that the learning algorithm can learn to produce correct solutions to the future state of the system without having to observe any labeled data. We develop a convolutional kernel which encodes the constraint that must be satisfied by any steady-state solution to the heat flow problem, and we use this kernel to determine the loss function. By seeking to minimize this loss, the network learns to satisfy the differential equations of heat transport, effectively learning the underlying physics of the system despite never explicitly being shown the outcomes for any given initial condition.

While the physical system examined throughout this paper is 2-D heat transport, the methods developed are extremely general and can be applied to any system defined by partial differential equations and theoretically capable of being solved by the finite difference method (even if it is not practical to do so).

This points us towards to possibility of encoding the equations we have discovered, which govern many physical environments of interest to us, into neural networks. It would be quite convenient if we could encode this information directly in our learning agents, such that they could benefit from the knowledge we've already gained about how the world works.

\section{Related Work} \label{related}
This work exists at the intersection of two different lines of research pursued by disparate research communities. The first line of work, weakly supervised machine learning, is pursued primarily within the machine learning community and seeks to reduce the data requirements of machine learning applications. It aims to do so by incorporating some form of prior knowledge, either to augment existing data, or to create context-aware learning algorithms that are able to achieve high performance with less data. The second line of work involves the use of physics informed machine learning for modeling physical systems, and is pursued primarily within the mathematical physics and engineering communities.

\cite{dawn17weak-supervision} defines weak supervision as a unified approach to incorporating various types of weak signal into the machine learning pipeline. These forms of weak signal include crowdsourced \cite{whitehill2009crowdsource}, noisy \cite{natarajan2013noisy}, or sparse labels (as in active and semi-supervised learning) \cite{zhu2009semi-supervised,settles2012active-learning}. Additionally, weak signal can be specified in the form of constraints and invariances over the output space (often provided by domain experts), or in terms of weak or biased classifiers (as in transfer learning or boosting). Although these techniques and approaches are disparate, they are unified by their aim to alleviate the need for vast quantities of data to solve machine learning problems.

The present work is most closely connected to work that aims to incorporate domain-specific prior knowledge in the form of constraints over the output space. The nearest cousin to this work is \cite{stewart2016label-free}, in which physical constraints are specified over the output trajectories that must be satisfied by solutions for problems in motion-tracking, enabling it to be done in a label-free way. We extend this work to the broader domain of prediction in physical systems governed by non-linear PDEs. In natural language processing, \cite{liang2013semantics,artzi2011bootstrap-semantics} seek to semantically parse statements or questions (i.e. convert them into their logical forms) with weak supervision signals (e.g. in the form of responses to queries rather than the meaning of the query itself). Recent works \cite{clarke2010semantic-response,guu2017language2programs} in this area apply constraints on the output space to provide weak signal to the semantic parser.

The second line of research that the present work extends leverages machine learning techniques to either discover the underlying dynamics of a system governed by unknown PDEs, or to build faster and more accurate differential equation solvers. \cite{raissi2017hidden,raissi2018deep-hp} use traditional machine learning techniques and deep learning techniques respectively for both predicting the future of time-dependent dynamical systems and for discovering their underlying equations. \cite{farimani2017transport} uses conditional generative models to find the equilibrium solution to a number of transport problems faster than traditional iterative methods. \cite{han2017overcoming-curse,sirigano2017dgm} leverage deep learning techniques to approximately solve partial differential equations is high-dimensional spaces, where traditional iterative techniques break down. All of these techniques use large quantities of data in a supervised way, unlike the present work.

\section{Background}\label{background}
\subsection{Heat-Transport}\label{heat-transport}
In 2-D heat transport, we consider a flat square plate made of some thermally conductive material that is insulated along its edges. Heat is applied to the plate in some way, and our goal is to model the way thermal energy moves through the plate. The initial condition is given by $T(x,y,0)$, and we wish to determine $T(x,y,t)$, the temperature field on the plate at time $t$. In our model we assume that non-zero elements of $T(x,y,0)$ represent an applied heat i.e. heat applied the point $(x,y)$ on the grid for the duration of the transport experiment. Under ideal assumptions, it can be shown that $T$ satisfies the two dimensional heat equation \cite{fourier1807heat}
$${\partial{T}\over\partial{t}} = c^2 \nabla^2T = c^2 ({\partial^2{T}\over \partial{x^2}} + {\partial^2{T} \over \partial{y^2}})$$
where $c > 0$ is a constant for the thermal conductivity of the plate. We can also study solutions that do not vary with time, known as the steady-state solutions to the system
$${\partial{T}\over\partial{t}} = 0$$
In this case we get the Laplace Equation:
\begin{equation} \label{steady-state}
\nabla^2T = {\partial^2{T}\over \partial{x^2}} + {\partial^2{T} \over \partial{y^2}} = 0
\end{equation}
Solutions to the Laplace equation are known as {\it harmonic functions}, and the particular solution to the steady-state heat transport problem is the harmonic function which also satisfies the initial condition of the system. When heat is applied only to the boundary of the plate, as it is in the cases we study in this paper, equation~\ref{steady-state} is known as the {\it Dirichlet boundary problem} \cite{dirichlet1850boundary}.

\subsection{Finite Difference}\label{finite-difference}
Finite difference is an iterative method used to compute exact solutions to partial differential equations via a discretization of the equations, and an update rule that is defined by the equations. To solve the 2D steady state heat equation using finite difference, we need to discretize $\nabla^2T$=0. Considering evenly spaced 2-D grid, the discretized form of $\nabla^2T$=0 for node ($i,$j) would be:
\begin{equation} \label{finite-difference-eqn}
T_{i,j} = {{T_{i+1,j}+T_{i-1,j}+T_{i,j+1}+T_{i,j-1}}\over 4}
\end{equation}
The nodal relation expressed in the above equation is solved iteratively by applying the rule above at each node (point in the grid) until convergence.

\section{Model}
\subsection{Approach}\label{approach}
The aim of this work is to train a fully convolutional neural network to directly infer the solution to the Laplace equation~(\ref{steady-state}) when given the initial condition as input (i.e. train the neural network to be a solver for the Dirichlet boundary problem \cite{dirichlet1850boundary}). We accomplish this without ever seeing solutions to the boundary problem by encoding the differential equations into a {\it physics-informed loss function}, described in section~\ref{physics-informed-kernel}, which motivates the network to find the solution without the use of supervision in the form of data. The architecture of the network is described in section~\ref{architecture}.

Each instantiation of the Dirichlet boundary problem is given to the neural network in the form of a $n \times n$ image matrix representing a thermally conductive plate, with the value in each entry representing the temperature applied to each point on the plate. In this work, we only apply heat to the boundary of the plate, so the only non-zero entries in the input are at the boundaries. Zero values in the input represent points on the plate with no temperature applied, meaning they can change over the course of time, as they are influenced by the temperature at neighboring points on the thermally conductive plate.

The desired output is also an $n \times n$ image matrix representing the temperature values at each point on the plate after the heat flow process has converged to an equilibrium temperature distribution.

The network is trained with randomly initialized boundary conditions. Because the network is never provided solutions to the boundary value problem, we can generate new data points on the fly virtually free of cost by creating new $n \times n$ matrices with the boundaries filled in by random values. In this work, we choose four random temperatures uniformly between 0 and 100 and set the temperatures of the top, right, left and bottom of the plate to be constant, and equal to each of the random temperatures correspondingly. An example input and output are shown in Figure~\ref{network-architecture}.

In section~\ref{progressive}, we describe how the output image is downsampled repeatedly and {\it multiscale loss function} is computed. Downsampling and training over multiple losses in this fashion dramatically speeds up training and improves the quality of the output images.

\subsection{Network Architecture}\label{architecture}
\begin{figure}[H]
\vskip -0.2in
\begin{center}
\centerline{\includegraphics[width=\columnwidth]{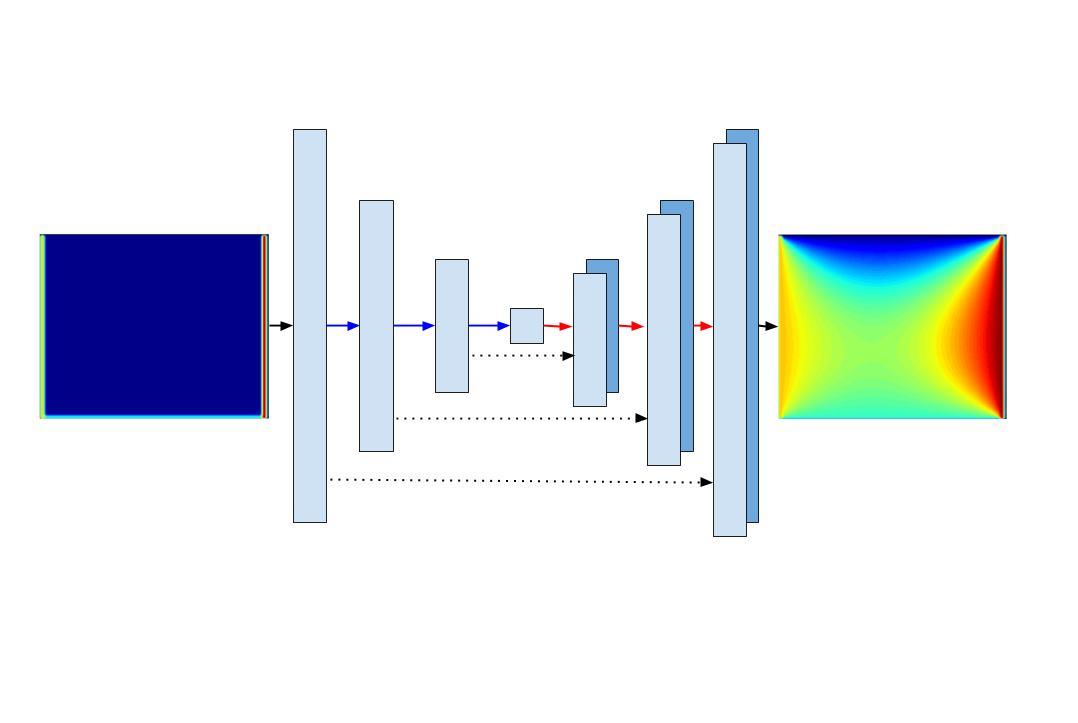}}
\vskip -0.6in
\caption{Encoder-decoder U-Net \cite{ronneberger2015unet} architecture of the network. Each encoding layer is connected to its corresponding decoder layer via a skip connection. The network is fully convolutional, and thus can be scaled to arbitrary size by adding layer. The input is the initial condition, and the output is the equilibrium condition.}
\label{network-architecture}
\end{center}
\end{figure}

The architecture of the network is a fully convolutional encoder-decoder network adapted from the U-Net architecture \cite{ronneberger2015unet}. The fully convolutional network is comprised of several encoding convolutional layers that decrease the image size to that of the latent space (in our case, $512 \times 1 \times 1$). The decoding layers consist of transposed convolutions on the output of the previous layer concatenated with the corresponding encoding layer. The concatenations amount to skip connections from each encoding layer to its corresponding output layer. Ultimately, an image of the original input size is recovered, and the aim is for the final image to represent the equilibrium temperatures on the plate. The architecture is shown in Figure~\ref{network-architecture}.

The motivation for using a fully convolutional architecture is to flexibly use the same architecture to solve problems at multiple scales. The purpose of the skip connections are to pass the boundary values of the input to the output layers, so the network  is not forced to memorize the structure of the input in its bottleneck layers. The architecture mirrors that of \cite{farimani2017transport}, in which this network architecture was used to solve a variety of differential equations in a supervised way.

\subsection{Physics Informed Kernel}\label{physics-informed-kernel}
The equilibrium condition defined in Equation~(\ref{steady-state}) can be encoded in a simple rule: the temperature at each point on the plate (that is not initially driven by a heat source) should be the average of its neighbors. In fact, it is by iteration of this rule that the {\it finite-difference method} for solving partial differential equations typically solves this problem, as was shown in equation~\ref{finite-difference-eqn}.

Examining this condition, we find that it can easily be encoded into a 3x3 convolutional kernel as follows:
\begin{table}[h]
\centering
\label{kernel}
\begin{tabular}{|l|l|l|}
\hline
0 & -1 & 0 \\
\hline
-1 & +4 & -1 \\
\hline
0 & -1 & 0  \\
\hline
\end{tabular}
\end{table}

This kernel is run convolutionally across an image, and the outputs flattened and normed, to calculate the loss of the output:
\begin{equation}\label{compute-loss}
\sum_{i,j}\mathrm (\mathsf{Conv2d}(\mathrm{kernel}, \mathrm{output})_{ij})^2
\end{equation}

By minimizing this loss, the neural networks learns to identify harmonic functions that form solutions to the given Dirichlet boundary problem specified by the initial condition, since the boundary of the output is fixed to be equal to the initial condition

Although this kernel can easily be identified by the structure of the heat equations, we show in section~\ref{learning-kernel} that this kernel can easily be learned if labels are provided (in the form of correct solutions to the heat flow equations). Although this kernel is easy derive and justify for the case of heat transport, in principle a similar kernel can be found for any system whose dynamics are defined by partial differential equations. In general, given an update rule for finite difference equations, it is easy to encode this rule into a convolutional kernel.

\subsection{Progressive Downsampling for Growing Loss}\label{progressive}
The network can have difficulty learning to output the heat distribution that minimizes the loss function for very large input sizes, in part because outputting a constant on the entire field has a loss of 0, if we ignore the boundary. As the image size grows larger, the boundary values becomes a proportionally smaller fraction of the total image, reducing the proportion of loss contributed by outputting incorrect values for the points near the border of the plate. Thus, the network is less able to accurately fill in correct temperatures, especially towards the center of the image, instead opting to output constant temperature fields.

In order to solve this problem, we adopt a modified version of the strategy of progressively growing the output of the network to the final problem size, first introduced in \cite{karras2017progressive}. In our case, instead of progressively adding decoding layers, as in \cite{karras2017progressive}, we compute the loss on several downsampled versions of the output image, and weight their contribution to the overall loss according to a weight vector $$\lambda = [\lambda_1, \lambda_2, \ldots, \lambda_n]$$ where each $\lambda_i$ is the weight of the loss on the $i^{\mathrm{th}}$ downsampled version of the output. The loss for each individual downsampled output is computed in the same way as described in section~\ref{physics-informed-kernel}, but the weight vector can slowly be tuned from $\lambda = [0, 0, \ldots, 1]$ to $\lambda = [1, \ldots, 0, 0]$. Early in training, this motivates the network to output images which satisfy the overall higher level structure of the desired output. We slowly increase the weights associated with getting the finer grained details of the output correct.

In our training, we choose a downsample factor of 4, until the images are of size $32 \times 32$, the scale at which training works well even without a progressively downsampled loss function.

\begin{figure}[H]
\begin{center}
\centerline{\hspace{3em}\includegraphics[scale=0.28]{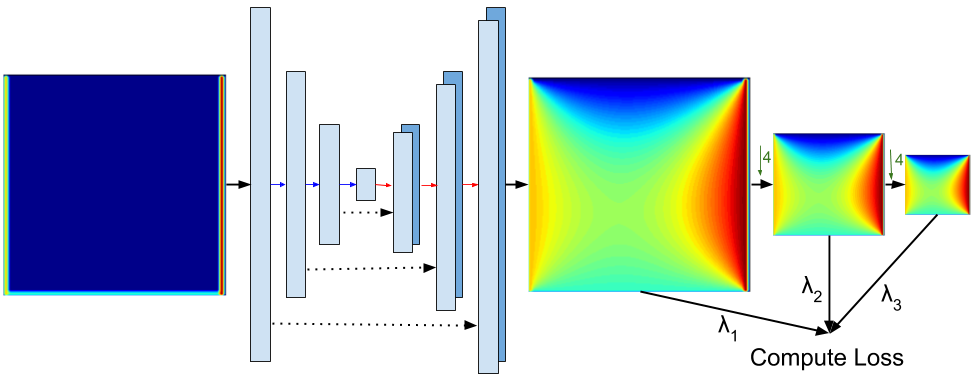}}
\caption{The output of the network is downsampled repeatedly by a factor of 4 until the dimension is 32 $\times$ 32. The downsample by $N$ operation used in this case simply uses ever $N^{\textrm{th}}$ pixel of the previous image. The loss in equation~\ref{compute-loss} is computed for each downsampled image, multipled by its corresponding weight $\lambda_i$ and added to compute the total loss.}
\label{loss-network}
\end{center}
\vskip -0.5in
\end{figure}

The general framework is shown in Figure~\ref{loss-network}

This framework can be viewed as a special case of curriculum learning \cite{bengio09curriculum}, where the difficulty of the curriculum scales as the weights in $\lambda$ become more concentrated on the largest scale. This provides a flexible framework by which output size can be used to craft a curriculum, and we suspect that it can be widely applied to multiscale problems in image generation, as well as other domains.

\section{Experiments \& Applications}

\subsection{Solving the Boundary Value Problem}\label{solving-bvp}
\begin{figure}
Initial Condition \ \ \ \ \ Correct Output \ \ \ \ \  Deep Learning
\begin{center}
\centerline{\includegraphics[width=\columnwidth]{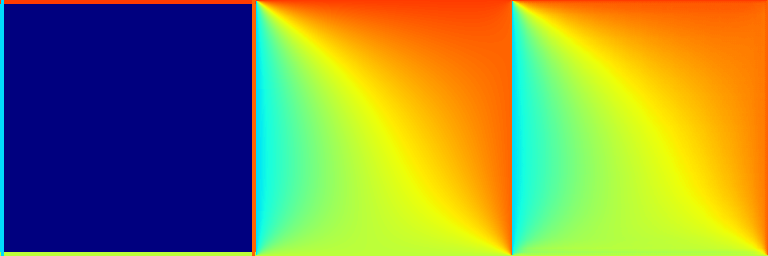}}
\centerline{\includegraphics[width=\columnwidth]{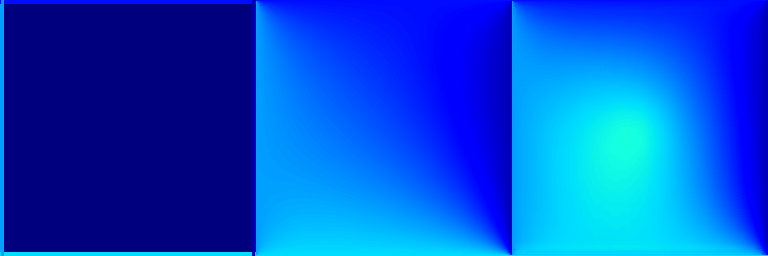}}
\centerline{\includegraphics[width=\columnwidth]{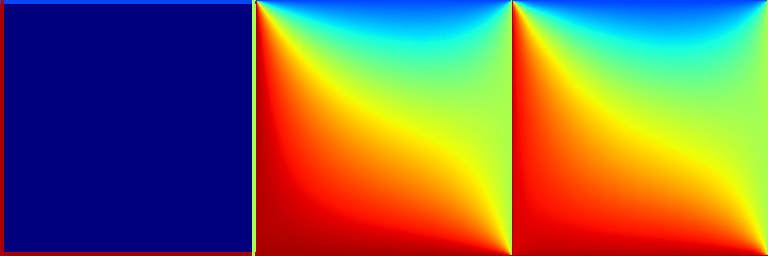}}
\centerline{\includegraphics[width=\columnwidth]{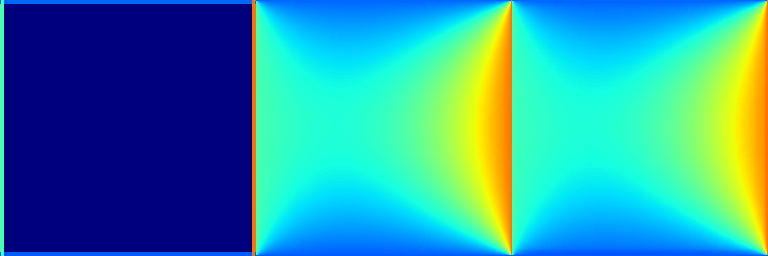}}
\centerline{\includegraphics[width=\columnwidth]{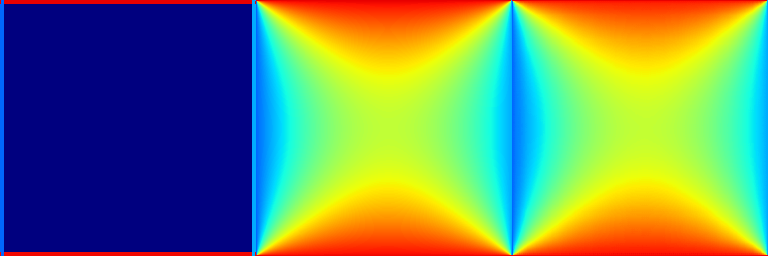}}
\caption{In the left column are the input initial conditions to the network. The second column shows the correct output temperature distribution (solved to high-precision via finite difference) that satisfies the equilibrium condition specified by equation~(\ref{steady-state}). The third column shows the output of the deep neural network that is trained to minimize (\ref{compute-loss}) These results are for inputs of size 256$\times$256}
 \label{results256}
\end{center}
\end{figure}

Test results are shown for inputs of size 256$\times$256 in Figure~\ref{results256}. During training of this experiment, the network was never shown the correct answer to the boundary value problem, but only ever given random initializations and trained to minimize the loss in equation(~\ref{compute-loss}). In the figure a number of images are shown containing the boundary value problem specified, the true equilibrium temperatures given that boundary condition (solved to high precision via finite difference), and the output of the neural network after training.

The network has sixteen hidden layers (8 encoding, 8 decoding) and has been trained for 128 epochs using a progressive downsampling strategy to grow the loss function as described in section~\ref{progressive}. The optimization is done with an Adam optimizer \cite{kingma2015adam}. Despite never seeing a correct desired output for the boundary value problem, the average per-pixel output error is only 1.39\% with a standard deviation of 1.24\%. The test examples are unseen during training, and are generated randomly according to the same procedure as during training.

\subsection{Solving at Very Large Scale}
As described in section~\ref{progressive}, the network has difficulty learning the correct solution at larger scales, because outputting constant values along the entire image becomes a viable strategy to achieve a low loss. However, by using the strategy of progressive downsampling to grow the loss function, the network is able to easily find solutions to the boundary value problem, achieving good results in just a few epochs. The basic method, without progressive downsampling, fails entirely to learn.

\begin{figure}
Initial Condition \ \ \ \ \ Correct Output \ \ \ \ \  Deep Learning
\begin{center}
\centerline{\includegraphics[width=\columnwidth]{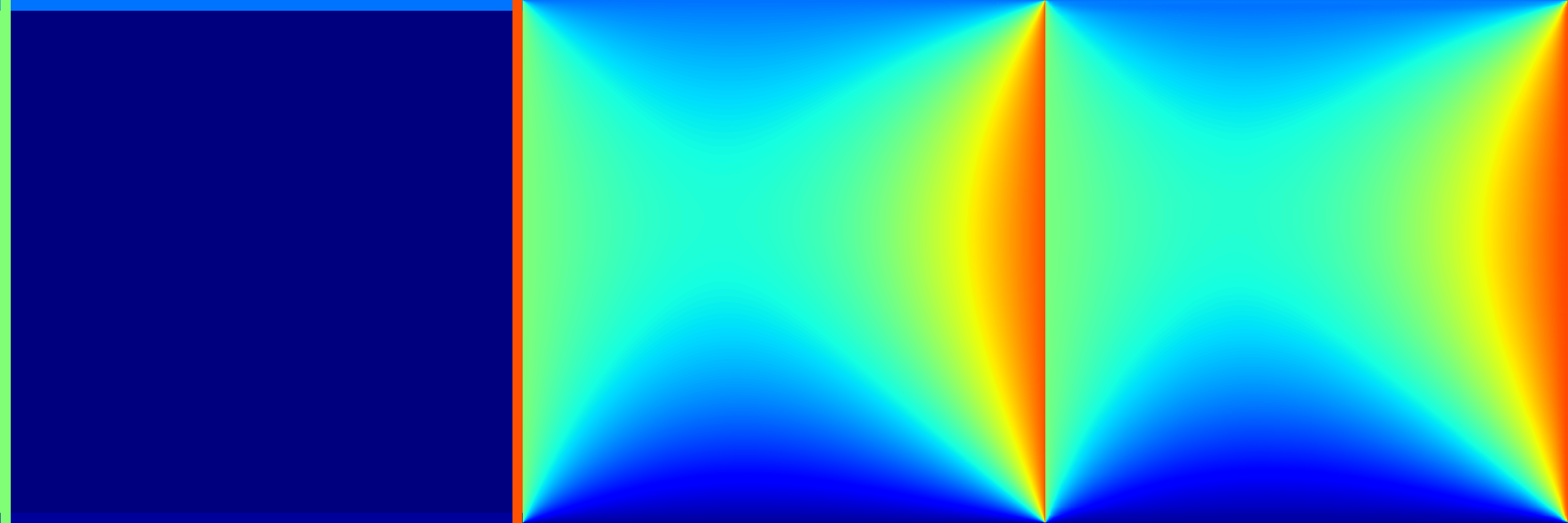}}
\centerline{\includegraphics[width=\columnwidth]{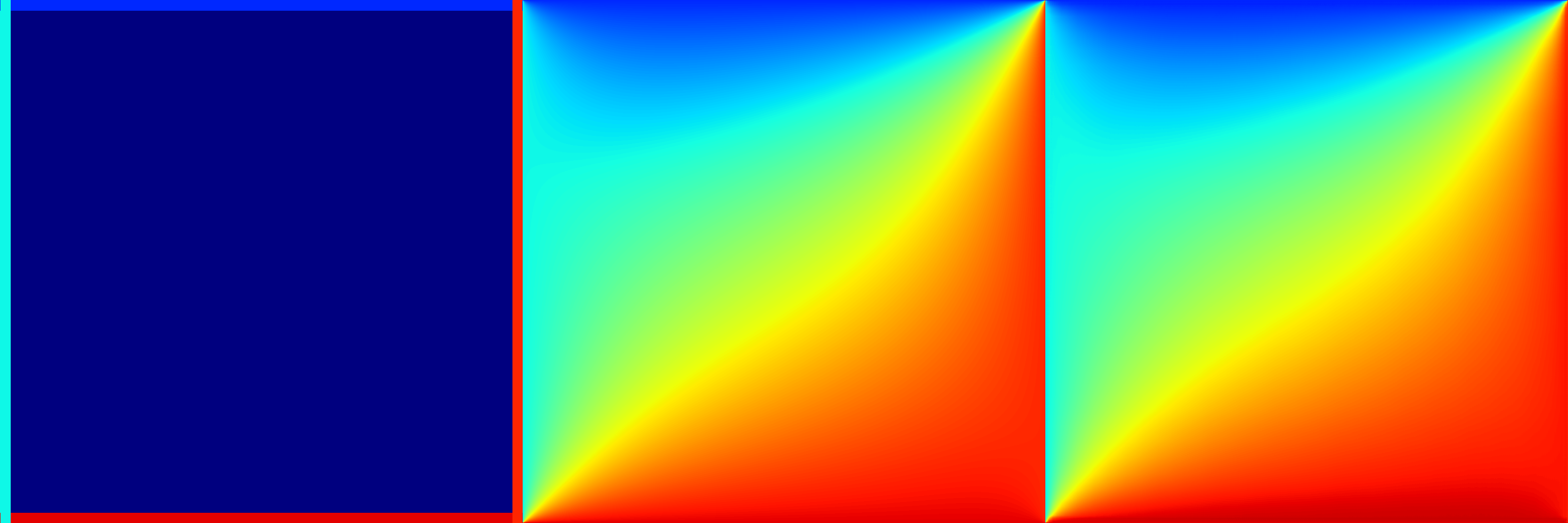}}
\centerline{\includegraphics[width=\columnwidth]{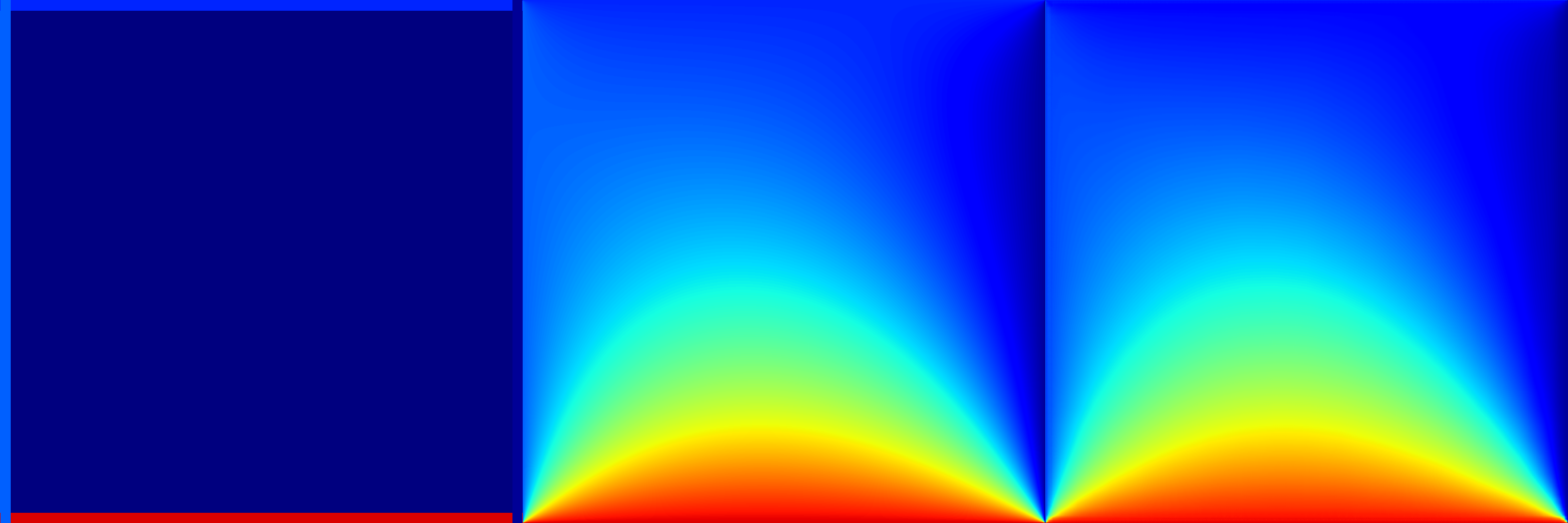}}
\centerline{\includegraphics[width=\columnwidth]{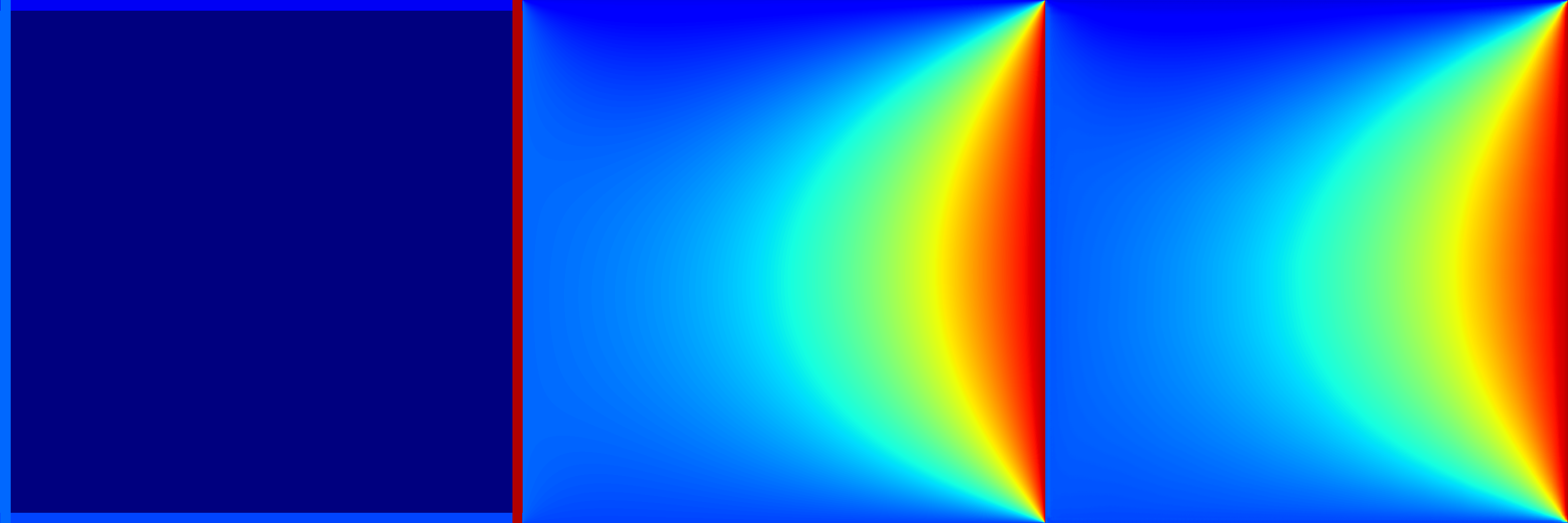}}
\centerline{\includegraphics[width=\columnwidth]{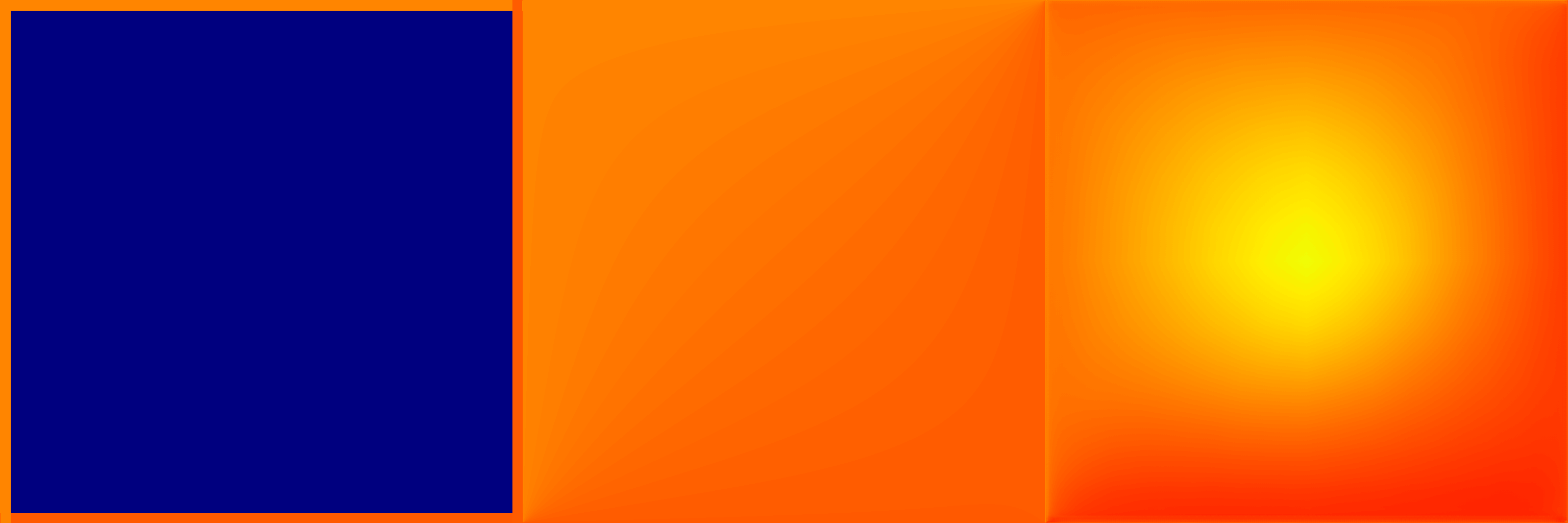}}
\caption{In the left column are the input initial conditions to the network. The second column shows the correct output temperature distribution (solved to high-precision via finite difference) that satisfies the equilibrium condition specified by equation~(\ref{steady-state}). The third column shows the output of the deep neural network that is trained to minimize \ref{compute-loss}. These results are for inputs of size 1024$\times$1024}
\label{results1024}
\end{center}
\end{figure}

Results for inputs of size 1024$\times$1024 are show in in Figure~\ref{results1024}. The network has twenty hidden layers (10 encoding, 10 decoding) and has been trained for 128 epochs, optimized with Adam \cite{kingma2015adam}. The average per-pixel output error is only 1.41\% with a standard deviation of 1.64\%. The test examples are unseen during training, and are generated randomly according to the same procedure as during training.

\subsection{Learning the Convolutional Kernel}\label{learning-kernel}
The kernel defined in section~\ref{kernel}, copied below, can be easily determined by inspection of the differential equations defining heat transport (i.e. the form of the Laplace equation~(\ref{steady-state})).
\begin{table}[h]
\centering
\begin{tabular}{|l|l|l|}
\hline
0 & -1 & 0 \\
\hline
-1 & +4 & -1 \\
\hline
0 & -1 & 0  \\
\hline
\end{tabular}
\end{table}

This may not be the case in general, however, as the local invariants satisfied by a system that obeys a certain set of differential equations may not be readily apparent. Furthermore, for some systems the differential equations governing their evolution may not even be known.

In this experiment, we show that for the case of heat transport, this convolutional kernel can be learned to high-precision as long as we have data. This data should consist of the equilibrium conditions of heat transport (i.e. solutions to the Dirichlet problem defined in equation~(\ref{steady-state}) for various initial conditions --- the initial conditions themselves are not required).

We generate data on the fly by randomizing the initial condition and running an iterative finite-difference solver to solve. The data generated is 8$\times$8 images which are converged to high precision. We then run a {\it learnable convolutional kernel} across the data and sum the absolute value of the outputs. The convolutional kernel is optimized via Adam, and its parameters seek to minimize the absolute value of the outputs of the convolution. After several thousand iterations on randomly generated data points, this procedure learns the following kernel:
\begin{table}[H]
\centering
\begin{tabular}{|l|l|l|}
\hline
0 & 0.0545 & 0.0001 \\
\hline
0.0545 & -0.2181 & 0.0545 \\
\hline
0 & 0.0547 & 0  \\
\hline
\end{tabular}
\end{table}
This is almost exactly a scaled version of the original kernel, which encodes the same rule: that each point on the eqilibrium heat distribution should be the average of the temperature of its four neighboring points. This shows that if we have data which describes converged conditions of the phenomenon of interest, we can learn the local rules that define equilibrium condition directly from this data. In principle, this can be used to discover differential equations governing the dynamics for unknown systems, although more work is needed to show that this can be done with more complex systems than heat-transport (e.g. systems with higher-order non-linearities).

\subsection{Application: Speeding Up Finite Difference Calculation} \label{application}
\begin{figure}
\begin{center}
\includegraphics[scale=0.5]{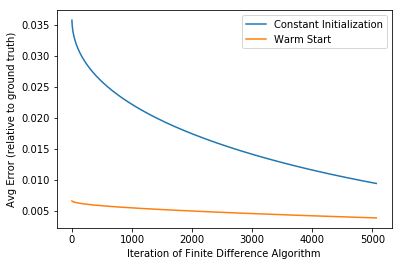}
\includegraphics[scale=0.5]{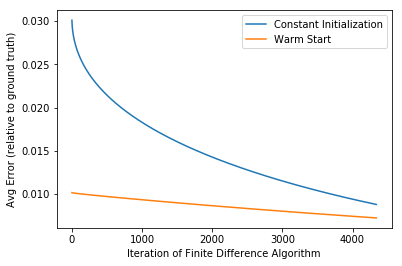}
\includegraphics[scale=0.5]{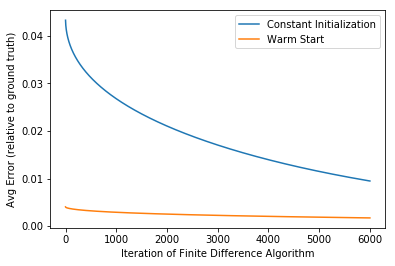}
\end{center}
\caption{A comparison of two strategies to solving finite difference are shown. The yellow line shows the average per-pixel error of the computed solution at each iteration of the finite difference algorithm when it is given the output of our trained neural network (size 256$\times$256 as in section~\ref{solving-bvp}) as an initialization. The blue line shows the average per-pixel error of the computed solution at each iteration of the finite difference algorithm when it is initialized with constant values equal to the average temperature of each of the borders of the plate. Average per-pixel error is computed relative to ``ground truth'', which is determined by running finite difference to a very high precision.}
\label{speed-up}
\end{figure}

The Finite Difference method described in Section~\ref{finite-difference} can be slow to converge to the correct solution \cite{remani2012numerical}, and is particularly sensitive to the initial condition given to the algorithm. Thus, we can use a trained neural network to provide an output that is approximately correct (within 1.5\% error), and this answer can be used as an initial condition to be refined by the finite difference method. The finite difference method can then compute the exact solution to a desired level of precision. Once trained, the forward pass through a neural network is fast, accounting for a negligible fraction of the time finite-difference algorithm takes to converge. Our results demonstrate that this ``warm-start'' approach for solving the heat equations converges much faster than a constant initialization --- i.e. one in which all pixels not on the border of the image are set to the average of pixels on the border. In Figure~\ref{speed-up}, we compare the errors from ground truth at each iteration of finite difference for the two different solvers, one that is given the output of the trained neural net as a warm start and one that is not. We compute ``ground truth'' by running finite difference to a very high level of precision relative to both initializations of the algorithm.

A more appropriate comparison for this method would be to hierarchical finite difference solvers, which are more often used in practice, but it is hard to make a direct comparison of the error between a hierarchical solver and our proposed method, as the hierarchical solver does not solve the full problem until near the end of its computation. In terms of wall-clock time, our method is not competitive with the hierarchical solvers, however it is possible that if a separately trained neural network is used at each layer of the hierarchy, our strategy will be superior to hierarchical finte difference algorithms.

\section{Conclusion}
We have shown how the equilibrium solution of the heat transport problem can be learned in a weakly supervised way by use of a physics informed loss function that encodes the local rules defined by the differential equations of heat transport. Further, we have shown that it is possible to use this technique to speed up finite difference calculations relative to the naive approach. We have also demonstrated that the local rule defining heat transport can be learned, in the form of a kernel, directly from data. In principle, the differential equations for heat transport can be replaced with those for any other phenomenon whose dynamics are defined by partial differential equations. This work points toward the possibility of encoding our knowledge of the dynamics of the world into neural networks, allowing them to learn how physical systems work even in the absence of data.



\bibliography{example_paper}

\newcommand{\etalchar}[1]{$^{#1}$}
\begin{thebibliography}{WfWB{\etalchar{+}}09}

\bibitem[AZ11]{artzi2011bootstrap-semantics}
Yoav Artzi and Luke Zettlemoyer.
\newblock Bootstrapping semantic parsers from conversations.
\newblock In {\em Proceedings of the Conference on Empirical Methods in Natural
  Language Processing}, EMNLP '11, pages 421--432, Stroudsburg, PA, USA, 2011.
  Association for Computational Linguistics.

\bibitem[BLCW09]{bengio09curriculum}
Yoshua Bengio, J{\'{e}}r{\^{o}}me Louradour, Ronan Collobert, and Jason Weston.
\newblock Curriculum learning.
\newblock In {\em Proceedings of the 26th Annual International Conference on
  Machine Learning, {ICML} 2009, Montreal, Quebec, Canada, June 14-18, 2009},
  pages 41--48, 2009.

\bibitem[CGCR10]{clarke2010semantic-response}
James Clarke, Dan Goldwasser, Ming-Wei Chang, and Dan Roth.
\newblock Driving semantic parsing from the world's response.
\newblock In {\em Proceedings of the Fourteenth Conference on Computational
  Natural Language Learning}, CoNLL '10, pages 18--27, Stroudsburg, PA, USA,
  2010. Association for Computational Linguistics.

\bibitem[Dir50]{dirichlet1850boundary}
P.G.L Dirichlet.
\newblock {\em \"{U}ber einen neuen Ausdruck zur Bestimmung der Dictigkeit
  einer unendlich d\"{u}nnen Kugelschale, wenn der Werth des Potentials
  derselben in jedem Punkte ihrer Oberfl\"{a}che gegeben ist}.
\newblock Abh K\"{o}ngilich. Preuss. Akad. Wiss., 1850.

\bibitem[FGP17]{farimani2017transport}
Amir~Barati Farimani, Joseph Gomes, and Vijay~S. Pande.
\newblock Deep learning the physics of transport phenomena.
\newblock {\em CoRR}, abs/1709.02432, 2017.

\bibitem[Fou07]{fourier1807heat}
Jean-Baptiste~Joseph Fourier.
\newblock 1807.

\bibitem[GPLL17]{guu2017language2programs}
Kelvin Guu, Panupong Pasupat, Evan~Zheran Liu, and Percy Liang.
\newblock From language to programs: Bridging reinforcement learning and
  maximum marginal likelihood.
\newblock {\em CoRR}, abs/1704.07926, 2017.

\bibitem[GrMH13]{graves2013speech}
A.~Graves, A.~r.~Mohamed, and G.~Hinton.
\newblock Speech recognition with deep recurrent neural networks.
\newblock In {\em 2013 IEEE International Conference on Acoustics, Speech and
  Signal Processing}, pages 6645--6649, May 2013.

\bibitem[HJE17]{han2017overcoming-curse}
Jiequn Han, Arnulf Jentzen, and Weinan E.
\newblock Overcoming the curse of dimensionality: Solving high-dimensional
  partial differential equations using deep learning.
\newblock {\em CoRR}, abs/1707.02568, 2017.

\bibitem[KALL17]{karras2017progressive}
Tero Karras, Timo Aila, Samuli Laine, and Jaakko Lehtinen.
\newblock Progressive growing of gans for improved quality, stability, and
  variation.
\newblock {\em CoRR}, abs/1710.10196, 2017.

\bibitem[KB14]{kingma2015adam}
Diederik~P. Kingma and Jimmy Ba.
\newblock Adam: {A} method for stochastic optimization.
\newblock {\em CoRR}, abs/1412.6980, 2014.

\bibitem[LJK13]{liang2013semantics}
Percy Liang, Michael~I. Jordan, and Dan Klein.
\newblock Learning dependency-based compositional semantics.
\newblock {\em Comput. Linguist.}, 39(2):389--446, June 2013.

\bibitem[MKS{\etalchar{+}}15]{mnih2015atari}
Volodymyr Mnih, Koray Kavukcuoglu, David Silver, Andrei~A. Rusu, Joel Veness,
  Marc~G. Bellemare, Alex Graves, Martin Riedmiller, Andreas~K. Fidjeland,
  Georg Ostrovski, Stig Petersen, Charles Beattie, Amir Sadik, Ioannis
  Antonoglou, Helen King, Dharshan Kumaran, Daan Wierstra, Shane Legg, and
  Demis Hassabis.
\newblock Human-level control through deep reinforcement learning.
\newblock {\em Nature}, 518:529 EP --, Feb 2015.

\bibitem[NDRT13]{natarajan2013noisy}
Nagarajan Natarajan, Inderjit~S Dhillon, Pradeep~K Ravikumar, and Ambuj Tewari.
\newblock Learning with noisy labels.
\newblock In C.~J.~C. Burges, L.~Bottou, M.~Welling, Z.~Ghahramani, and K.~Q.
  Weinberger, editors, {\em Advances in Neural Information Processing Systems
  26}, pages 1196--1204. Curran Associates, Inc., 2013.

\bibitem[Rai18]{raissi2018deep-hp}
Maziar Raissi.
\newblock Deep hidden physics models: Deep learning of nonlinear partial
  differential equations.
\newblock {\em CoRR}, abs/1801.06637, 2018.

\bibitem[RBVR17]{dawn17weak-supervision}
Alex Ratner, Stephen Bach, Paroma Varma, and Chris Ré.
\newblock Weak supervision: The new programming paradigm for machine learning
  (blog post), 2017.

\bibitem[RDS{\etalchar{+}}15]{russakovsky2015imagenet}
Olga Russakovsky, Jia Deng, Hao Su, Jonathan Krause, Sanjeev Satheesh, Sean Ma,
  Zhiheng Huang, Andrej Karpathy, Aditya Khosla, Michael Bernstein,
  Alexander~C. Berg, and Li~Fei-Fei.
\newblock Imagenet large scale visual recognition challenge.
\newblock {\em Int. J. Comput. Vision}, 115(3):211--252, December 2015.

\bibitem[Rem12]{remani2012numerical}
Courtney Remani.
\newblock Numerical methods for solving systems of nonlinear equations, 2012.

\bibitem[RFB15]{ronneberger2015unet}
Olaf Ronneberger, Philipp Fischer, and Thomas Brox.
\newblock U-net: Convolutional networks for biomedical image segmentation.
\newblock In Nassir Navab, Joachim Hornegger, William~M. Wells, and
  Alejandro~F. Frangi, editors, {\em Medical Image Computing and
  Computer-Assisted Intervention -- MICCAI 2015}, pages 234--241, Cham, 2015.
  Springer International Publishing.

\bibitem[RK17]{raissi2017hidden}
Maziar Raissi and George~Em Karniadakis.
\newblock Hidden physics models: Machine learning of nonlinear partial
  differential equations.
\newblock {\em Journal of Computational Physics}, 2017.

\bibitem[SB98]{suttonbarto98RL}
Richard~S. Sutton and Andrew~G. Barto.
\newblock {\em Introduction to Reinforcement Learning}.
\newblock MIT Press, Cambridge, MA, USA, 1st edition, 1998.

\bibitem[SE16]{stewart2016label-free}
Russell Stewart and Stefano Ermon.
\newblock Label-free supervision of neural networks with physics and domain
  knowledge.
\newblock {\em CoRR}, abs/1609.05566, 2016.

\bibitem[Set12]{settles2012active-learning}
Burr Settles.
\newblock Active learning.
\newblock {\em Synthesis Lectures on Artificial Intelligence and Machine
  Learning}, 6(1):1--114, 2012.

\bibitem[SS17]{sirigano2017dgm}
J.~{Sirignano} and K.~{Spiliopoulos}.
\newblock {DGM: A deep learning algorithm for solving partial differential
  equations}.
\newblock {\em ArXiv e-prints}, August 2017.

\bibitem[SSS{\etalchar{+}}17]{silver2017alphazero}
David Silver, Julian Schrittwieser, Karen Simonyan, Ioannis Antonoglou, Aja
  Huang, Arthur Guez, Thomas Hubert, Lucas Baker, Matthew Lai, Adrian Bolton,
  Yutian Chen, Timothy Lillicrap, Fan Hui, Laurent Sifre, George van~den
  Driessche, Thore Graepel, and Demis Hassabis.
\newblock Mastering the game of go without human knowledge.
\newblock {\em Nature}, 550:354 EP --, Oct 2017.
\newblock Article.

\bibitem[SVL14]{sutskever2014seq2seq}
Ilya Sutskever, Oriol Vinyals, and Quoc~V Le.
\newblock Sequence to sequence learning with neural networks.
\newblock In Z.~Ghahramani, M.~Welling, C.~Cortes, N.~D. Lawrence, and K.~Q.
  Weinberger, editors, {\em Advances in Neural Information Processing Systems
  27}, pages 3104--3112. Curran Associates, Inc., 2014.

\bibitem[WfWB{\etalchar{+}}09]{whitehill2009crowdsource}
Jacob Whitehill, Ting fan Wu, Jacob Bergsma, Javier~R. Movellan, and Paul~L.
  Ruvolo.
\newblock Whose vote should count more: Optimal integration of labels from
  labelers of unknown expertise.
\newblock In Y.~Bengio, D.~Schuurmans, J.~D. Lafferty, C.~K.~I. Williams, and
  A.~Culotta, editors, {\em Advances in Neural Information Processing Systems
  22}, pages 2035--2043. Curran Associates, Inc., 2009.

\bibitem[ZG09]{zhu2009semi-supervised}
Xiaojin Zhu and Andrew~B. Goldberg.
\newblock Introduction to semi-supervised learning.
\newblock {\em Synthesis Lectures on Artificial Intelligence and Machine
  Learning}, 3(1):1--130, 2009.

\end{thebibliography}
\bibliographystyle{alpha}
\end{document}